\documentclass{bmvc2k}

\usepackage{array}
\usepackage{multirow}
\usepackage{url}

\title{Pandora: Description of a Painting Database for Art Movement Recognition with Baselines and Perspectives}

 \addauthor{Corneliu Florea}{corneliu.florea@upb.ro;}{1}
 \addauthor{R\u{a}zvan Condorovici}{rcondorovici@imag.pub.ro;}{1}
 \addauthor{Constantin Vertan}{constantin.vertan@upb.ro;}{1}
 \addauthor{Raluca Boia}{rboia@imag.pub.ro;}{1}
 \addauthor{Laura Florea}{laura.florea@upb.ro;}{1}
 \addauthor{Ruxandra Vr\^{a}nceanu}{rvranceanu@imag.pub.ro;}{1}

\addinstitution{
 Image Processing and Analysis Laboratory\\
 University ``Politehnica'' of Bucharest\\
 Bucharest, Romania
}

\runninghead{Florea et al. }{Pandora: Description of a Painting Database}

\begin{document}

\maketitle

\begin{abstract}
To facilitate computer analysis of visual art, in the form of paintings, we introduce Pandora
(Paintings Dataset for Recognizing the Art movement) database, a collection of digitized paintings
labelled with respect to the artistic movement. Noting that the set of databases available as
benchmarks for evaluation is highly reduced and most existing ones are limited in variability and
number of images, we propose a novel large scale dataset of digital paintings. The database
consists of more than 7700 images from 12 art movements. Each genre is illustrated by a number of
images varying from 250 to nearly 1000. We investigate how local and global features  and
classification systems are able to recognize the art movement. Our experimental results suggest
that accurate recognition is achievable by a combination of various categories.
\end{abstract}


\section{Introduction}
The remarkable expansion of the digital data during the last period favored a much easier access to
works of art for the general public. Great efforts were put lately into creating automatic image
processing solutions that facilitate a better understanding of art \cite{Cornelis2011}. These
solutions may aim at obtaining high-quality and high-fidelity digital versions of paintings
\cite{Martinez2002} or may address various aspects such as: image diagnostics, virtual restoration,
color rejuvenation etc. as discussed in the review of Stork et al. \cite{Stork2009}. Another more
appropriate to the ultimate goal of computers is the context recognition. One of the broadest
possible implementation of context recognition is the automatic art movement identification.

According to Artyfactory \cite{Artyfactory}, \emph{art movements} are  ``collective titles that are
given to artworks which share the same artistic ideals, style, technical approach or timeframe''.
While some works are clearly set into a single art movement, others are in the transition period,
as painters loved to experiment new ideas, leading to creation of a new movement. Also while the
actual characteristics place a work in some art movement, its author, for personal reasons, refused
to be categorized in such a way, giving birth to disputes.

In this paper, we look into the problem of computational  categorization of digitized paintings
into artistic genres (or art movements). In contrast to other directions of image classification,
such as scene or object recognition, where large databases and evaluation protocols do exist, such
an aspect is less emphasized for digitized paintings. Typically, the evaluation of a new method is
carried on a small database with few paintings belonging to few genres. Given the latest advances
of machine learning, two aspects should be noted: (1) deep networks with the many parameters easily
overfit on small databases and (2) to have progress, we need larger databases.

In this paper we start by reviewing painting collections introduced  in prior art and we follow by
describing the proposed database. Next, to form a baseline, we continue by reporting the
performance of various popular image descriptors and machine learning systems on the introduced
database. The paper ends with discussions and conclusions.

\section{Related work}

In the last period multiple solutions issued automatic analysis of  visual art and especially
paintings using computer vision techniques. However, most of the research is based on
medium--to--small databases. A summary of such methods is presented in table
\ref{Tab:StateOfTheArt}. One may easily note the size of the databases (and implicitly the number
of art movements investigated) increased with time, while the reported performance decreased until
it stabilized in the range of 50-70\% for correct art movement recognition. Some of the most
representative databases used for art movement identification are:
\begin{itemize}

\item Artistic genre dataset \cite{Gunsel2005}. Images, gathered from Web Museum-Paris, were set in
the following art movements: Classicism, Cubism, Impressionism, Surrealism, Expressionism.

\item Artistic genre dataset \cite{Zujovic2009}. Images from various Internet sources were
categorized into 5 genres : Abstract, Impressionism, Cubism, Pop Art and Realism.

\item Painting genre dataset \cite{Siddiquie2009}: Images collected from the Internet were grouped
into: Abstract expressionist, Baroque, Cubist, Graffiti, Impressionist and Renaissance.

\item Artistic style dataset \cite{Shamir2010}: Paintings from 9 painters were grouped intro three
art movements: Impressionism, Abstract expressionism and Surrealism.

\item Artistic genre dataset \cite{Arora2012} with images collected from Artchive fine-art dataset
and grouped into: fine-art genres: Renaissance, Baroque, Impressionism, Cubism, Abstract,
Expressionism and Pop art.

\item Paintings-91 dataset \cite{Khan2014} with images collected from the Internet. While the
database is larger than the previous ones, only paintings corresponding to painters that have the
majority of works into one art movement got a genre label. It resulted in a smaller database
illustrating Abstract expressionism, Baroque, Constructivism, Cubism, Impressionism, Neo-classical,
Pop art, Post-impressionism, Realism, Renaissance, Romanticism, Surrealism and Symbolism. Probably
this is the most structured database previously proposed.

\item Artistic genre dataset \cite{Condorovici2015} is the basis of the proposed database. We
increased that dataset by adding more images to illustrate the existing art movements and added 4
new ones.

\item Artistic genre dataset \cite{Agarwal2015} contains images collected from WikiArt and grouped
into: Abstract-expressionism, Baroque, Cubism, Impressionism, Expressionism, Pop Art, Rococo,
Realism, Renaissance and Surrealism.

\end{itemize}

Concluding, many of the databases previously used, are small and contain non-standard evaluation
protocols allowing overfitting. Thus, a larger scale database with fixed evaluation protocol should
be beneficial for further development on the topic.

{\setlength\tabcolsep{4pt}
\begin{table}[t]
\begin{center}
    \caption{Art movement recognition solutions with the size of used databases. The database size refers only
     to the database used for art movement recognition, as in some cases larger databases have been
     implied for other purposes. The value for recognition rate (RR) is the one reported by the respective
     work while the ``test ratio'' is the percentage used for testing from the overall database (CV-stands for cross validation).
    We kindly ask the reader to retrieve all the details from the respective work.}
    \label{Tab:StateOfTheArt}
    \begin{tabular}{|c|c| c|c|c|}
        \hline
        \textbf{Method} & 
         \begin{tabular}{c} \textbf{Move-} \\ \textbf{ments}\end{tabular} & 
         \begin{tabular}{c} \textbf{Db.} \\ \textbf{size}\end{tabular} & 
            \textbf{RR.} & 
            \textbf{Test ratio} 
           \\ \hline
        Gunsel et al. \cite{Gunsel2005}         & 3 & 107 & 91.66\% & 53.5\% \\ \hline
        Zujovic et al. \cite{Zujovic2009}       & 5 & 353 & 68.3\%  & 10\% CV \\ \hline
        Siddiquie et al. \cite{Siddiquie2009}   & 6 & 498 & 82.4\%  & 20\% CV \\ \hline
        Shamir et al. \cite{Shamir2010}                & 3 & 517 & 91\%    & 29.8\% \\ \hline
        Arora\&ElGammal\cite{Arora2012}      & 7 & 490 & 65.4\%  & 20\% CV\\ \hline
        Khan et al. \cite{Khan2014}             & 13 & 2338 & 62.2\% & 46.53\% CV \\ \hline
        Condorovici et al.\cite{Condorovici2015}& 8 & 4119 & 72.24\% & 10\% CV \\ \hline
        Agarwal  et al. \cite{Agarwal2015}      & 10 & 3000 & 62.37\% & 10\% CV \\ \hline \hline
        \emph{Proposed}                                & \emph{12}& \emph{7724}& \emph{54.7\%}  & \emph{25\% CV}  \\ \hline

    \end{tabular}
\end{center}
\end{table}
}

\section{Pandora database}

Our main contribution is the creation of a new and extensive dataset of art images\footnote{The
up-to-date database with pre-computed features data reported here is available at
\url{http://imag.pub.ro/pandora/pandora_download.html} }. While we follow the Paintings-91 database
\cite{Khan2014}, our dataset is significantly larger, it was build around art movements and not
painters  and we tried to span wider time periods from antiquity to current periods. The later
property should help the automatic study of style evolution, of thematic evolution and cross-time
relationship identifications.

\begin{figure*}
\centering
\begin{tabular}{c}
    \begin{tabular}{cccc}
       \includegraphics[width=0.19\textwidth]{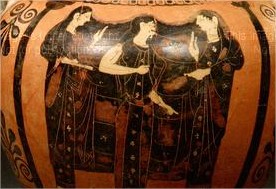} &
        \includegraphics[width=0.185\textwidth]{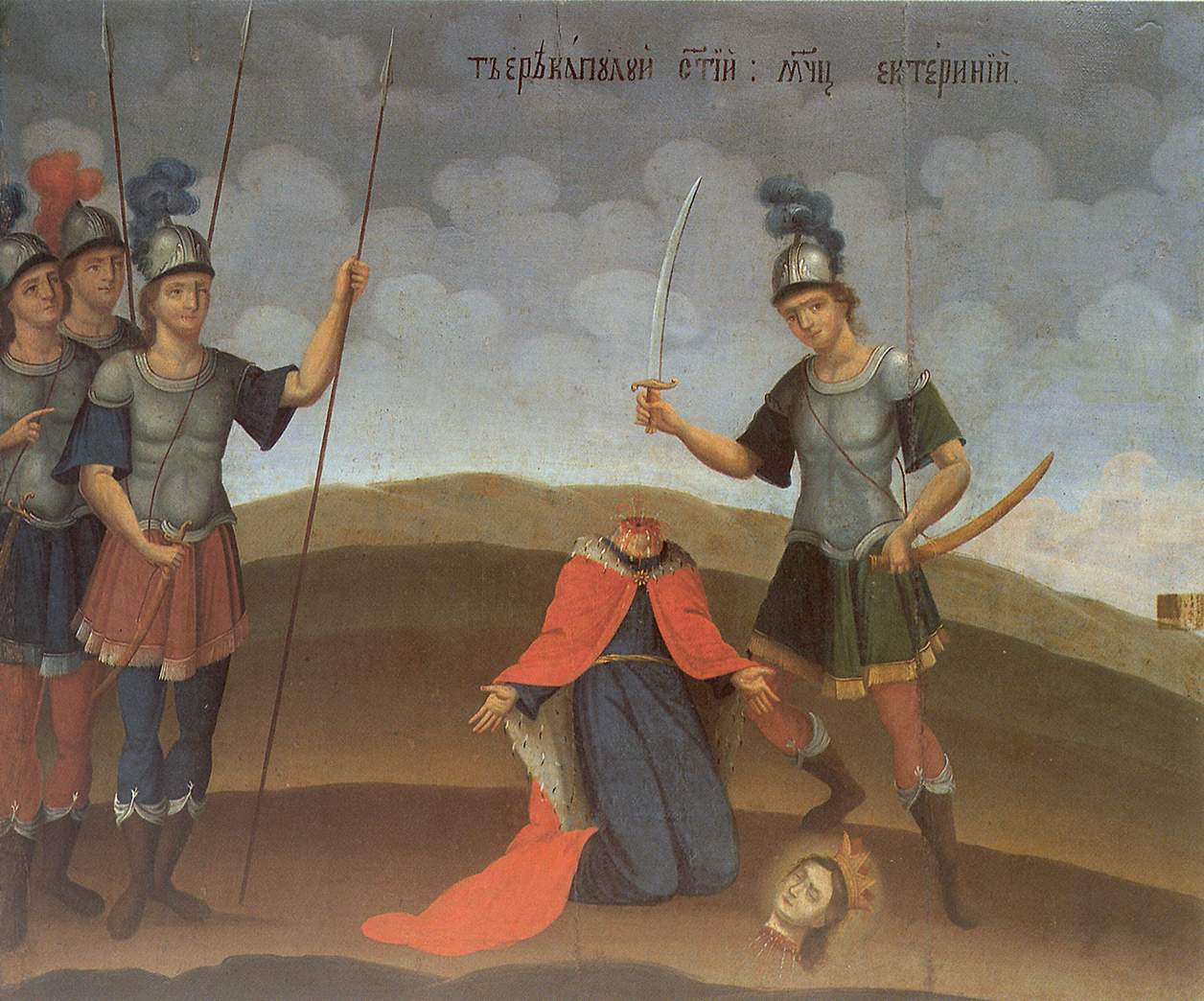}&
        \includegraphics[width=0.175\textwidth]{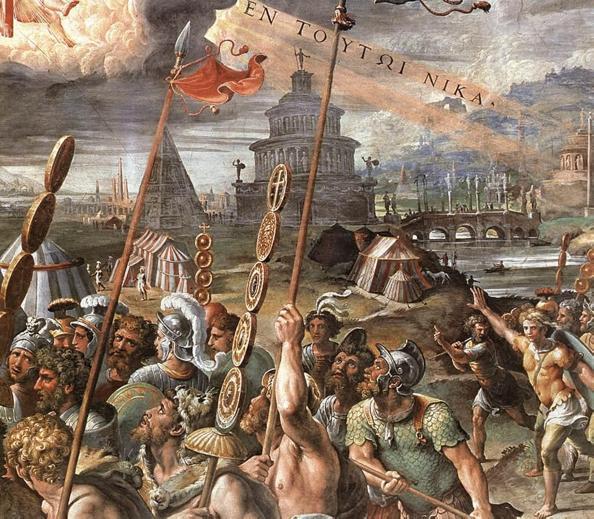} &
        \includegraphics[width=0.19\textwidth]{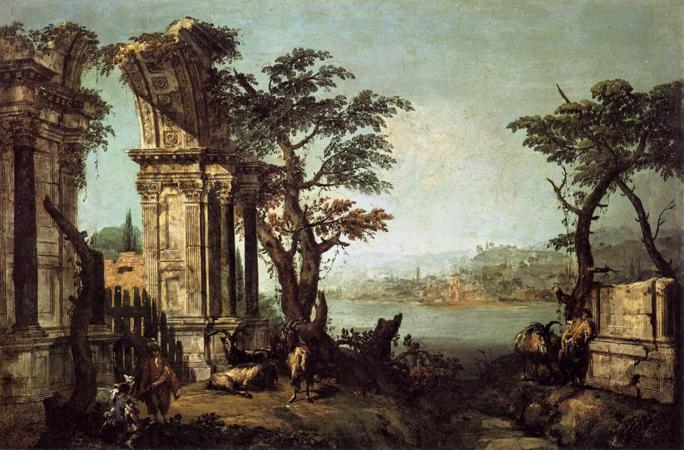}  \\
        Old Greek Pottery & Orthodox Iconoclasm & High Renaissance & Baroque
    \end{tabular} \\

    \begin{tabular}{cccc}
        \includegraphics[width=0.19\textwidth]{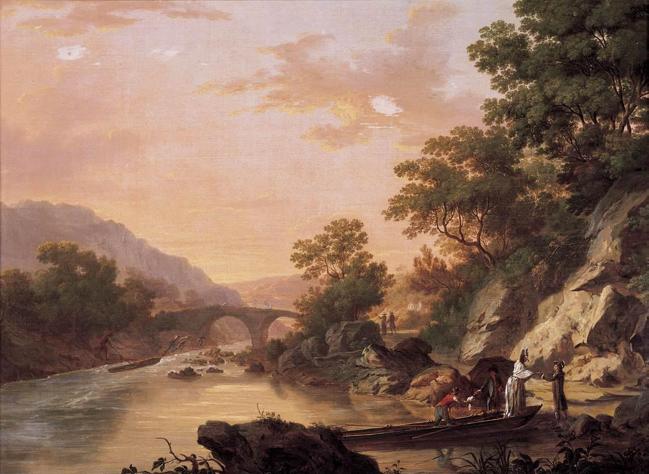} &
        \includegraphics[width=0.19\textwidth]{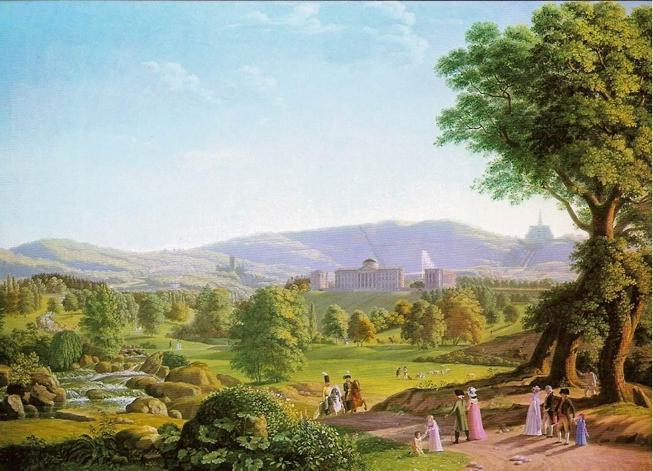} &
        \includegraphics[width=0.19\textwidth]{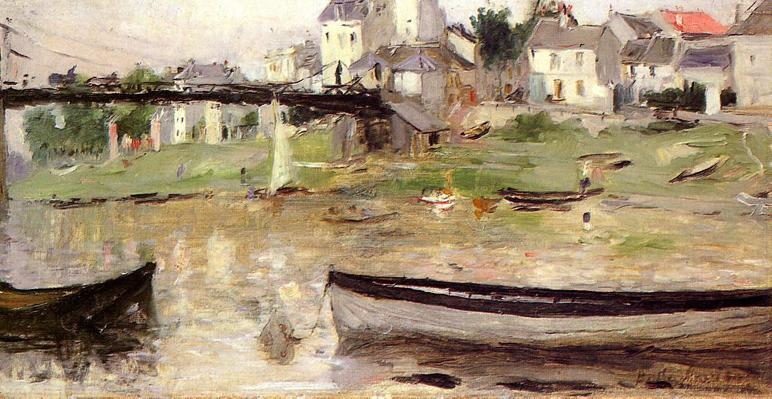}&
        \includegraphics[width=0.19\textwidth]{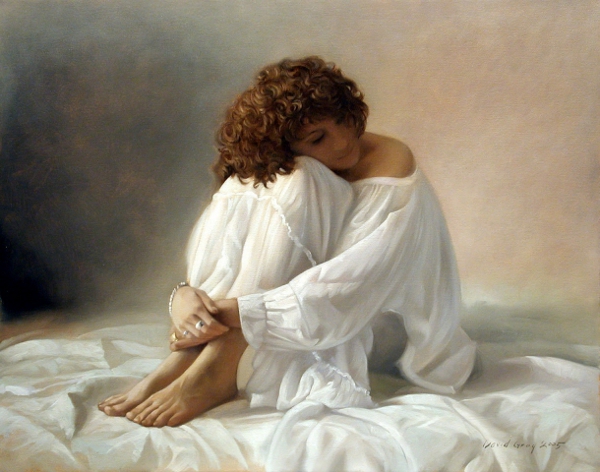} \\
        Rococo & Romanticism &  Impressionism &  Realism
    \end{tabular} \\

    \begin{tabular}{cccc}
        \includegraphics[width=0.19\textwidth]{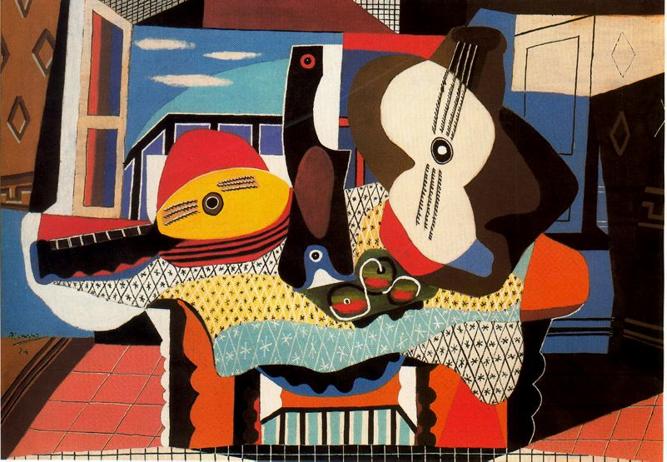} &
        \includegraphics[width=0.185\textwidth]{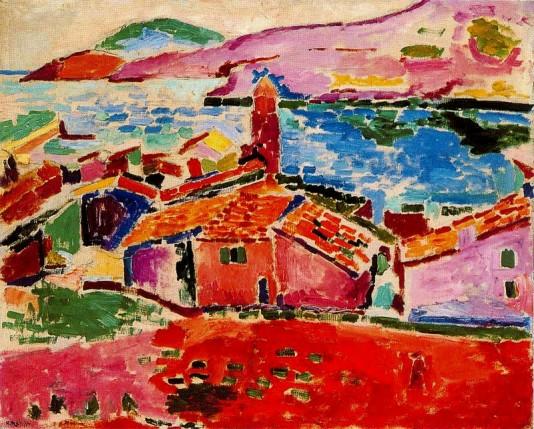}&
        \includegraphics[width=0.185\textwidth]{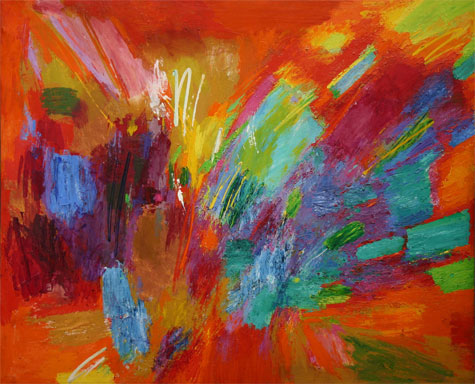}&
        \includegraphics[width=0.185\textwidth]{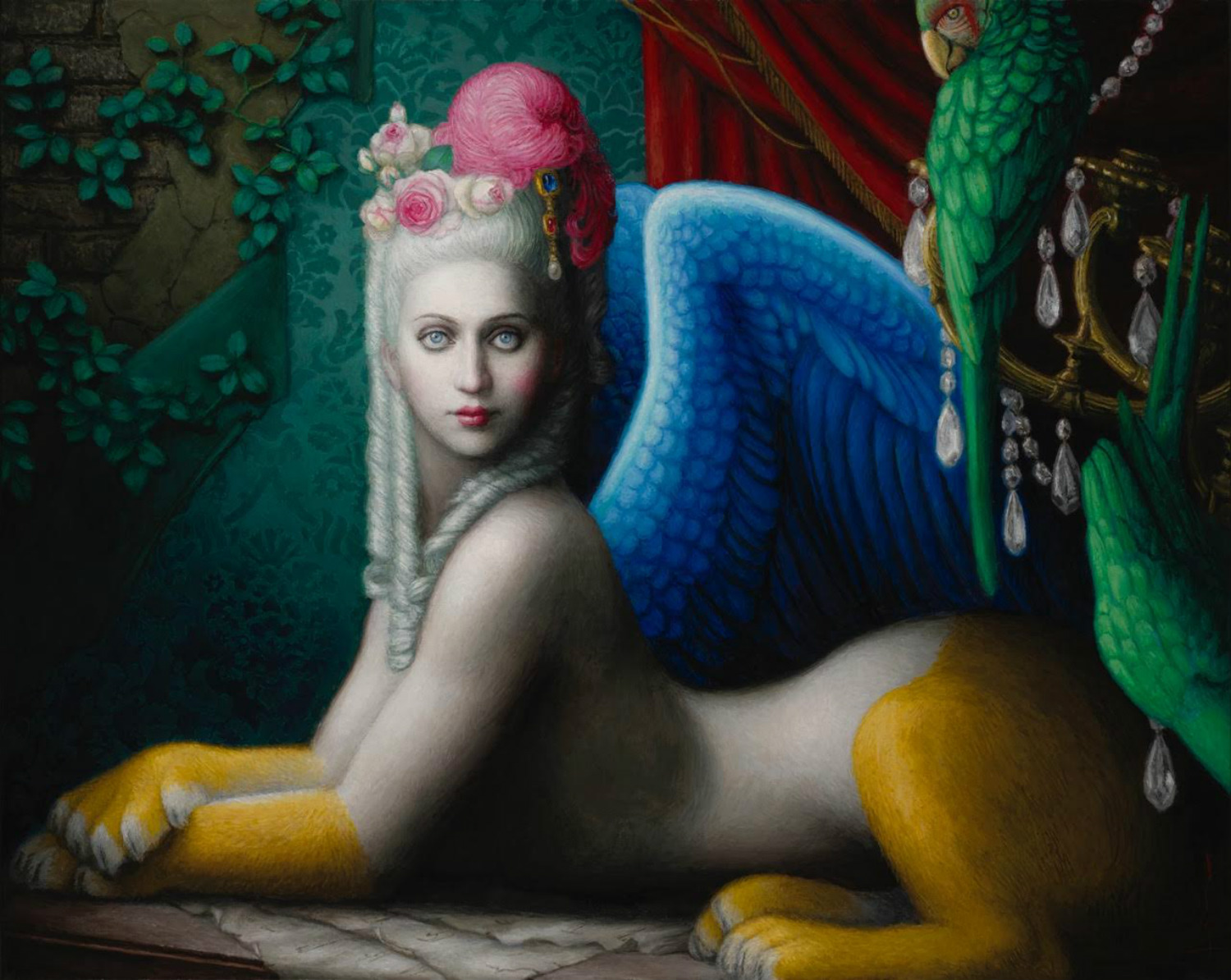} \\
        Cubism & Fauvism & Abstract &   Surrealism
    \end{tabular} \\
\end{tabular}
\caption{The 12 art movements illustrated in the proposed database. \label{Fig:Painting_Examples} }
\end{figure*}

The Pandora (Paintings Dataset for Recognizing the Art movement) dataset consists of 7724 images
from 12 movements: old Greek pottery, iconoclasm, high renaissance, baroque, rococo, romanticism,
impressionism, realism, cubism, fauvism, abstract-expressionism and surrealism. The precise
database structure is shown in table \ref{Tab:PandoraStucture} and some examples representative for
the art movements are in figure \ref{Fig:Painting_Examples}. We kindly ask the reader to note some
of difficulties in distinguishing between genres: the main difference between Abstract and Fauvism
is the less \emph{natural order} in the structure of the Abstract works, while the Fauvism tends
``to use color to express joy``. Baroque has a darker tone with respect to Romanticism while the
later depicts "exotism or extraordinary things" . The difference between Realism and Surrealism is
that the later illustrate ``irrational juxtaposition of images'' \cite{Artyfactory} (e.g. such as
wings attached to the girl). Yet thinking in computer terms, to detect the \emph{irrational} of
\emph{joy} in an image is extremely hard. Thus we consider that to achieve such goals, one needs,
first, an appropriate database of considerable size and variability.

\begin{table}[t]
\begin{center}
    \caption{The structure of the Pandora database. \label{Tab:PandoraStucture}}
    \begin{tabular} {|c|c|c|}
    \hline
    \textbf{Art movement} & \textbf{No. of paintings} & \textbf{Historical period} \\ \hline
    Old Greek pottery &  350        & Antiquity \\ \hline
    Iconoclasm        &  665        & Middle age \\ \hline
    High renaissance  &  812        & 1490 - 1527 \\ \hline
    Baroque           &  960        & 1590 - 1725 \\ \hline
    Rococo            &  844        & 1650 - 1850 \\ \hline
    Romanticism       &  874        & 1770 - 1850 \\ \hline
    Impressionism     &  984        & 1860 - 1925 \\ \hline
    Realism           &  307        & 1848 - present \\ \hline
    Cubism            &  920        & 1900 - present \\ \hline
    Abstract-expressionism & 340    & 1920 - present \\ \hline
    Fauvism           & 426         & 1900 - 1950 \\ \hline
    Surrealism        & 242         & 1900 - present \\ \hline
    \end{tabular}
\end{center}
\end{table}

\section{Art movement recognition performance}

\subsection{Training and testing}
To separate the database training and testing parts, a 4-fold cross validation scheme was
implemented. The division into 4 folds exists at the level of each art movement, thus each image
being uniquely allocated into a fold. The same division was used for all further tests and it is
part of the database.

\subsection{Features and classifiers}

As ``there is no fixed rule that determines what constitutes an art movement'' and ''the artists
associated with one movement may adhere to strict guiding principles, whereas those who belong to
another may have little in common'' \cite{Artyfactory}, there cannot be a single set of descriptors
that are able to separate any two art movements.

Following the observations from prior works \cite{Arora2012}, \cite{Khan2014}, multiple categories
of feature descriptors should be used. For instance, to differentiate between impressionism and
previous styles, one of the main difference is the brush stroke, thus \emph{texture}. Old Pottery
and Orthodox Iconoclasm are older and use a limited \emph{color palette}. Also, one needs to
understand the content of the painting to distinguish between realism and surrealism (for
instance); thus, global \emph{composition} descriptor should be used.

To provide a baseline for further evaluation, we have tested various combinations of popular
feature extractors and classification algorithms.

The texture feature extractors used are :
\begin{itemize}
    \item \textbf{Histogram of oriented gradients} (HOG)  \cite{Dalal2005} which computes the
        oriented gradient in each pixel and accumulates the weight of each orientation into a
        histogram. It has been previously used in painting analysis \cite{Khan2014}, \cite{Agarwal2015}.
    \item \textbf{Pyramidal HOG} (pHOG) the above mentioned HOG is
    implemented on 4 levels of a Gaussian pyramid.
    \item \textbf{Color HOG} - the above mentioned HOG descriptor applied on each color plane of the RGB
    color space.
    \item \textbf{Local Binary Pattern }(LPB) \cite{Ojala:2002} is a histogram of quantized
    binary patterns pooled in a local image neighborhood of $3\times 3$ and restrained to a total of 58
    quantized non-uniform patterns. The LPB was used in painting
    description \cite{Khan2014}, \cite{Agarwal2015}.

    \item \textbf{Pyramidal LBP} (pLBP) - the above mentioned descriptor computed over 4 levels
    of a Gaussian pyramid.
    \item \textbf{Local Invariant Order Pattern} \cite{Wang:2011} - assume the order after sorting
    in the increasing intensity local samples.

    For HOG, LBP and LIOP we have relied on the implementation from the VLFeat library
    \cite{Vedaldi:2010}.
    \item \textbf{Edge Histogram Descriptor} (EHD) is part of the MPEG-7 standard. It accounts for the
    distribution of four basic gradient orientations within regular image parts. The implementation
    is based on BilVideo-7 library \cite{Bacstan2010}.

    \item \textbf{The spatial envelope, GIST} \cite{Oliva2001} describes the spatial character or
    shape of the painting and was previously used for painting categorization \cite{Agarwal2015}.

\end{itemize}

The color descriptors tested are:
\begin{itemize}

    \item \textbf{Discriminative Color Names} (DCN) \cite{Khan2013} - represents the dominant color
    retrieved through an information oriented approach. Here, we have used
    author provided code. The baseline form (Color Name) was successfully used to determine
    the style and the painter \cite{Khan2014}.

    \item \textbf{Color Structure Descriptor} (CSD) \cite{Manjunath:2001}, which is based on  color
    structure histogram, a generalization of the color histogram. The CSD accounts
    for some spatial coherence in the gross distribution of quantized colors within the image and  it
    has been shown that is able to differentiate between various art movements
    \cite{Huang2014}. We computed a 64 long CSD vector using the BilVideo-7 library \cite{Bacstan2010}.

\end{itemize}

Machine learning classification systems tested are:
\begin{itemize}

    \item \textbf{Support Vector Machine}. We have relied for its implementation on the Lib-SVM
    \cite{Chang:2011}. We used on the radial basis function c-SVM and followed, for each case, the
    optimization (i.e. exhaustive search in (cost,gamma) space) recommended by the LibSVM creators.

    \item \textbf{Random forest} \cite{breiman2001}. We have used 100 trees and unlimited depth. At
    each node we randomly look for a split in $N_1 =\sqrt{N}$ dimensions where $N$ is the input feature
    dimension.

    Let us note that before the development of the deep networks the
    random forests and support vector machines have been found to be the most robust families of
    classifiers \cite{fernandez2014}. Also, for small and diverse databases SVM and RF out-compete
    deep networks.

    \item \textbf{k-Nearest neighbor} (kNN). We have implemented 1-NN, 3-NN and 7-NN based on Euclidean
    distance. While we report the results in terms of correct recognition rate, the nearest neighbor results will give
    an indication about the retrieval performance as it may be translated in terms of precision--recall.
\end{itemize}
Furthermore we have tested several systems that were previously used for art movement recognition.
Inspired from previous work \cite{Arora2012}, we have run the Bag of Words (BoW) over SIFT keypoint
detector with a vocabulary of 500. We have also tested a combination of color description, texture
analysis based on Gabor filters and scene composition based on Gestalt frameworks
\cite{Condorovici2015}.

Additionally, while the database is small for such a purpose and thus not really suited for deep
learning, to have an indication of baseline performance, we have trained and evaluated a version of
Deep Convolutional Neural Network (CNN). Our implementation is based on the MatConvNet
\cite{Vedaldi:2015} library and LeNet architecture \cite{Lecun1998}.

\subsection{Results}
We report first the results achieved when various combinations of features and classifiers are used
(to be followed in table \ref{Tab:BasicFeaturePerf}). We also report, in tables \ref{Tab:confMats},
\ref{Tab:confMats2}, the confusion matrices for the best combination in each category: pLPB+SVM,
GIST+RF, CSD+SVM and respectively pLBP+CSD+SVM.

\begin{table}[tb]
\centering \caption{Recognition rates when various combinations of features and classifiers are
used on the Pandora database. We marked with bold the best
performance.\label{Tab:BasicFeaturePerf}}

\begin{tabular}{|c| c|c| c|c|c|}
    \hline
      Feat. / Class. & Random Forest & SVM & 1-NN  & 3-NN  & 7-NN \\ \hline
       HOG                 &   0.266     & 0.248 & 0.200 & 0.214 & 0.233 \\ \hline
       pHOG                &   0.342     & 0.364 & 0.262 & 0.266 & 0.267 \\ \hline
       colorHOG            &   0.268     & 0.277 & 0.213 & 0.221 & 0.236 \\ \hline
       LBP                 &   0.386     & 0.395 & 0.303 & 0.298 & 0.320 \\ \hline
       pLBP                &   0.459     & 0.525 & 0.368 & 0.362 & 0.377 \\ \hline
       LIOP                &   0.344     & 0.362 & 0.246 & 0.252 & 0.260 \\ \hline
       EHD                 &   0.319     & 0.287 & 0.270 & 0.267 & 0.286 \\ \hline\hline
       GIST                &   0.379     & 0.337 & 0.297 & 0.280 & 0.282 \\ \hline \hline
       DCN                 &   0.298     & 0.264 & 0.192 & 0.201 & 0.215 \\ \hline
       CSD                 &   0.435     & 0.489 & 0.337 & 0.3357 & 0.363 \\ \hline \hline
       pLBP + DCN          &   0.488     & 0.521  & 0.278 & 0.282 & 0.297 \\ \hline \hline
       pLBP + CSD          &   0.540     & \textbf{0.547} & 0.377 & 0.282 & 0.297 \\ \hline
\end{tabular}
\end{table}

{\setlength\tabcolsep{2pt}
\begin{table*}[t]
\centering
 \caption{Confusion Matrices for the best performers on each category. The art movement used acronyms
  are: old Greek pottery--\emph{Gre}, iconoclasm--\emph{Ico}, high renaissance--\emph{Ren},
  baroque--\emph{Bar}, rococo--\emph{Roc}, romanticism--\emph{Rom}, impressionism--\emph{Impr},
  realism--\emph{Real}, cubism--\emph{cub}, fauvism--\emph{Fauv}, abstract-expressionism--\emph{Abs}
  and surrealism--\emph{Sur}.
\label{Tab:confMats}}
\begin{tabular}{c}
    \begin{tabular}{|c| c|c|c|c| c|c|c|c| c|c|c|c|} 
    \hline
     & Gre & Ren & Icon & Roc & Rom & Bar & Impr & Cub & Abs & Fauv & Real & Sur \\ \hline
    Gre &   \textbf{200}  & 20   & 28   &  2   &  3   &  5   & 35   & 51   &  3   &  1   &  0   &  2 \\ \hline
    Ren   &   5   &\textbf{360}   & 24   & 54   & 47   &153   & 95   & 70   &  1   &  2   &  0   &  1 \\ \hline
    Icon &    7   & 21   &\textbf{559}   &  7   &  1   & 15   & 23   & 30   &  2   &  0   &  0   &  0 \\ \hline
    Roc &   4   & 90   & 21   &\textbf{272}   &129   &214   & 73   & 38   &  1   &  0   &  0   &  2 \\ \hline
    Rom &    9   &106   & 16   &165   &\textbf{219}   &183   &107   & 55   &  2   &  5   &  3   &  4 \\ \hline
    Bar  &  4   &173   & 21   &186   &121   &\textbf{292}   & 87   & 65   &  2   &  3   &  0   &  6 \\ \hline
    Impr &   10   & 97   & 47   & 66   &106   & 95   &\textbf{429}   &109   &  8   &  6   &  2   &  9 \\ \hline
    Cub &   14   & 93   & 65   & 42   & 55   & 76   &167   &\textbf{361}   & 12   & 12   &  2   & 21 \\ \hline
    Abs &   11   & 43   & 43   &  5   & 18   & 11   & 66   & 67   & \textbf{46}   &  2   &  0   & 28 \\ \hline
    Fauv&    7   & 53   & 20   & 15   & 35   & 43   &118   & 89   &  6   & \textbf{18}   &  1   & 21 \\ \hline
    Real&    2   & 10   & 16   & 22   & 25   & 17   & 39   & 16   &  1   &  3   &\textbf{135}   & 21 \\ \hline
    Surr&    2   & 13   & 16   &  7   & 16   &  7   & 43   & 37   & 14   &  7   &  2   & \textbf{78} \\ \hline
    \end{tabular} \\
     GIST + RF \\

    \begin{tabular}{|c| c|c|c|c| c|c|c|c| c|c|c|c|}
    \hline
         & Gre & Ren & Icon & Roc & Rom & Bar & Impr & Cub & Abs & Fauv & Real & Sur \\ \hline
    Gre & \textbf{339}   &  4   &  1   &  1   &  0   &  3   &  0   &  1   &  0   &  0   &  1   &  0 \\ \hline
    Ren   &    0   &\textbf{404}   & 16   & 49   & 66   &219   & 24   & 31   &  0   &  2   &  1   &  0 \\ \hline
    Icon &    0   &  8   &\textbf{576}   &  3   &  0   &  6   & 30   & 31   &  0   & 11   &  0   &  0 \\ \hline
    Roc &    0   &118   &  2   &\textbf{230}   &153   &262   & 51   & 18   &  0   &  1   &  8   &  1 \\ \hline
    Rom &     0   & 91   &  4   &113   &\textbf{322}   &228   & 74   & 31   &  0   &  1   & 10   &  0 \\ \hline
    Bar  &   0   &178   &  8   &152   &150   &\textbf{405}   & 42   & 18   &  0   &  1   &  5   &  1 \\ \hline
    Impr &    0   & 24   & 58   & 33   & 35   & 66   &\textbf{584}   &133   & 13   & 29   &  9   &  0 \\ \hline
    Cub &    2   & 43   & 42   & 38   & 21   & 32   &168   &\textbf{508}   & 16   & 30   &  5   & 15 \\ \hline
    Abs &    0   &  3   & 11   &  1   &  7   &  2   & 55   & 83   &\textbf{125}   & 45   &  0   &  8 \\ \hline
    Fauv&    0   &  9  &  15   &  8   &  8   & 11   &104   & 85   & 34   &\textbf{145}   &  3   &  4 \\ \hline
    Real&   1   & 18   & 14   & 30   & 29   & 42   & 69   & 38   &  4   &  7   & \textbf{50}   &  5 \\ \hline
    Surr&   0   &  3   &  5   &  8   & 16   &  5   & 44   & 65   & 15   & 19   &  5   & \textbf{57} \\ \hline
    \end{tabular} \\

     CSD + SVM \\
    \end{tabular}
\end{table*}

\begin{table*}[t]
\centering
 \caption{Confusion Matrices for the best performers on each category. The art movement used acronyms
  are: old Greek pottery--\emph{Gre}, iconoclasm--\emph{Ico}, high renaissance--\emph{Ren},
  baroque--\emph{Bar}, rococo--\emph{Roc}, romanticism--\emph{Rom}, impressionism--\emph{Impr},
  realism--\emph{Real}, cubism--\emph{cub}, fauvism--\emph{Fauv}, abstract-expressionism--\emph{Abs}
  and surrealism--\emph{Sur}.
\label{Tab:confMats2}}
\begin{tabular}{c}
    \begin{tabular}{|c| c|c|c|c| c|c|c|c| c|c|c|c|}
    \hline
    & Gre & Ren & Icon & Roc & Rom & Bar & Impr & Cub & Abs & Fauv & Real & Sur \\ \hline
    Gre &    \textbf{271}   &  4   &  2   &  1   &  5   &  1   & 11   & 27   &  5   & 18   &  2   &  3 \\ \hline
    Ren   &     4   &\textbf{444}   &  1   & 40   & 60   &166   & 36   & 58   &  1   &  0   &  1   &  1\\ \hline
    Icon &   42   &  1   &\textbf{598}   &  1   &  0   &  3   &  3   & 10   &  3   &  4   &  0   &  0 \\ \hline
    Roco &    0   & 48   &  3   &\textbf{324}   &136   &232   & 63   & 36   &  0   &  0   &  1   &  1 \\ \hline
    Rom &     3   & 94   &  4   &139   &\textbf{339}   &182   & 80   & 27   &  1   &  3   &  0   &  2 \\ \hline
    Bar  &    1   &126   &  3   &165   &136   &\textbf{441}   & 47   & 37   &  0   &  2   &  0   &  2 \\ \hline
    Impr &   9   & 29   & 10   & 49   & 92   & 43   &\textbf{618}   &112   &  6   & 14   &  1   &  1 \\ \hline
    Cub &  63   & 29   & 23   & 17   & 29   & 28   &128   &\textbf{570}   &  8   & 17   &  0   &  8 \\ \hline
    Abs &  20   & 11   & 21   &  7   & 22   & 17   & 56   & 91   & \textbf{59}   & 23   &  2   & 11 \\ \hline
    Fauv&  33   &  5   &  5   &  1   & 18   &  9   &111   & 58   & 11   &\textbf{163}   &  2   & 10 \\ \hline
    Real&  15   & 11   &  4   &  7   & 17   & 17   & 33   & 23   & 13   & 18   &\textbf{141}   &  8 \\ \hline
    Surr&  18   & 13   & 10   &  4   & 23   & 19   & 17   & 40   & 13   & 13   &  4   & \textbf{68} \\ \hline
    \end{tabular} \\ pLBP + SVM \\ \\

    \begin{tabular}{|c| c|c|c|c| c|c|c|c| c|c|c|c|}
    \hline
    & Gre & Ren & Icon & Roc & Rom & Bar & Impr & Cub & Abs & Fauv & Real & Sur \\ \hline
    Gre &  \textbf{307}   &  0   &  0   &  0   &  0   &  0   &  2   & 40   &  0   &  0   &  1   &  0 \\ \hline
    Ren   &     0   &\textbf{473}   &  2   & 49   & 63   &159   & 24   & 41   &  0   &  1   &  0   &  0 \\ \hline
    Icon &    0   &  1   &\textbf{616}   &  1   &  1   &  2   & 10   & 31   &  1   &  2   &  0   &  0 \\ \hline
    Roc &    0   & 83   &  1   &\textbf{301}   &163   &214   & 59   & 23   &  0   &  0   &  0   &  0 \\ \hline
    Rom &     0   & 77   &  0   &104   &\textbf{372}   &184   & 74   & 63   &  0   &  0   &  0   &  0 \\ \hline
    Bar  &    0   &132   &  0   &169   &148   &\textbf{437}   & 50   & 24   &  0   &  0   &  0   &  0 \\ \hline
    Impr &   0   & 17   &  7   & 31   & 43   & 36   &\textbf{696}   &143   &  0   & 10   &  1   &  0 \\ \hline
    Cub &   2   & 21   & 13   & 11   & 21   & 17   &119   &\textbf{706}   &  1   &  9   &  0   &  0 \\ \hline
    Abs &   0   &  0   &  4   &  1   &  6   &  1   & 53   &200   & \textbf{62}   & 13   &  0   &  0 \\ \hline
    Fauv&  0   &  2   &  2   &  2   &  6   &  1   &126   &190   &  3   & \textbf{93}   &  1   &  0 \\ \hline
    Real&  0   &  3   &  0   &  9   & 13   & 11   & 56   & 79   &  0   &  4   &\textbf{131}   &  1 \\ \hline
        Surr& 0   &  1   &  3   &  4   &  7   &  2   & 33   &159   &  4   &  6   &  0   & \textbf{23} \\ \hline
    \end{tabular} \\
     pLBP + CSD + SVM  \\
    \end{tabular}
\end{table*}
}

Secondly we report comparatively the best performance of aggregated systems in table
\ref{Tab:SystemPerf}. We note that for this particular database, the best performance is achieved
by a standard combination of features (pyramidal LBP + Color Structure Descriptor) with a Support
Vector Machine.

While one may find disappointing the performance of various established systems, this is perfectly
explainable. For the Bag of Words there is too much variability between keypoints to find a common
ground; instead of the baseline version tested here, one should opt for much larger vocabularies
with accurate compression to keep memory requirements low. Regarding the performance of the
DeepCNN, the reported value should be perceived as a lower boundary, as the database is too small
for directly training nets with tens of thousands of variables, since no data augmentation was
implemented and the images being resized at $32\times 32$ lost some of the defining
characteristics.

\begin{table}[tb]
\centering \caption{Recognition rates when various systems are used. \label{Tab:SystemPerf}}

\begin{tabular}{|c| c|c| c|c|c|}
    \hline
      \textbf{System} & \textbf{Performance} \\ \hline
      pLBP + CSD +SVM   &   \textbf{0.547}   \\ \hline
       BoW                &   0.352           \\ \hline
       Condorovici et al. \cite{Condorovici2015} & 0.379 \\ \hline
       Deep CNN            &   0.486     \\ \hline
\end{tabular}
\end{table}

\section{Discussion and conclusions}

The best achieved performance was by a combination of pyramidal LBP and Color Structure Descriptor.
One may expect the addition of GIST to further increase the performance, but this does not happen,
probably due to the curse of dimensionality (the features dimension reaching 800); in such a case a
feature selection method should be used, but we consider it outside the scope of the current paper.

The next important observation is that different descriptors do a good job separating some currents
and not so good on identifying others. For instance, the CSD separates excellently the Orthodox
Iconoclasm which has a unique color palette (due to degradation in time and reduced colors
available at creation), but it is not able to separate Fauvism from Impressionism as both use the
same colors but distributed differently. The Surrealism is hard to separate by everything else
except GIST as it is the only tested feature able to describe the scene composition. Yet the GIST
is not able to distinguish the Fauvism from Impressionism as local texture makes the difference. In
contrast, the pLBP confusion between Fauvism and Impressionism is much reduced.

Overall, the confusion between Abstract and Cubism is large. As Cubism is defined by the
extraordinary apparition of straight lines, to address it, one should try to introduce features
appropriate to describe rectilinear objects.

Concluding we propose a new painting database annotated with art movements labels and divided in 4
folds to prepare it for rigorous evaluation. The database is significantly larger than the ones
previously used.  We have tested a multitude of popular features and classifiers and we have
identified the weak and strong points of each of them. We also suggest some directions for future
research that we anticipate to be beneficial for progress in the field.


\section*{Acknowledgment}

This work is supported by a grant of the Romanian National Authority for Scientific Research and
Innovation, CNCS – UEFISCDI,  number PN-II-RU-TE-2014-4-0733

\bibliography{pandoraDB}

\end{document}